\pdfoutput=1
\documentclass[10pt, conference]{IEEEtran}
\IEEEoverridecommandlockouts
\usepackage{cite}
\usepackage{amsmath,amssymb,amsfonts}
\usepackage{algorithm,algorithmic}
\usepackage{graphicx}
\usepackage{textcomp}
\usepackage{xcolor}
\usepackage{times}
\usepackage{url}
\usepackage{booktabs}
\usepackage{setspace}
\usepackage[caption=false]{subfig}

\usepackage[mathscr]{euscript}
\usepackage{enumitem}
\newcommand\eatpunct[1]{}
\usepackage{cleveref}
\usepackage{subfloat}
\usepackage{multirow}

\begin{document}

\title{Deep-HOSeq: Deep Higher Order Sequence Fusion for Multimodal Sentiment Analysis
\thanks{Jiwei Wang$^\ddagger$, Zhefeng Ge$^\ddagger$, Rujia Shen$^\ddagger$ contributed equally to this work and share the second authorship of this work.}
\thanks{Sunny Verma$^{\dagger,\mathsection}$ is the corresponding author of this work.}
}

\author{\IEEEauthorblockN{Sunny Verma$^{\dagger,\mathsection}$, Jiwei Wang$^\ddagger$, Zhefeng Ge$^\ddagger$, Rujia Shen$^\ddagger$, Fan Jin$^\ddagger$, Yang Wang$^{\dagger}$, Fang Chen$^{\dagger}$, and Wei Liu$^\dagger$}
\IEEEauthorblockA{$^{\dagger}$\textit{University of Technology Sydney, Sydney, Australia} \\
$^{\ddagger}$\textit{Hangzhou Dianzi University, Hangzhou, China}\\
\{sunny.verma, yang.wang, fang.chen, wei.liu\}@uts.edu.au\\
\{wangjiwei, gezhefeng, rujiashen, fanjin\}@hdu.edu.cn\\
}

}

\maketitle

\begin{abstract}
Multimodal sentiment analysis utilizes multiple heterogeneous modalities for sentiment classification. The recent multimodal fusion schemes customize LSTMs to discover intra-modal dynamics and design sophisticated attention mechanisms to discover the inter-modal dynamics from multimodal sequences. Although powerful, these schemes completely rely on attention mechanisms which is problematic due to two major drawbacks 1) deceptive attention masks, and 2) training dynamics. Nevertheless, strenuous efforts are required to optimize hyperparameters of these consolidate architectures, in particular their custom-designed LSTMs constrained by attention schemes. In this research, we first propose a common network to discover both intra-modal and inter-modal dynamics by utilizing basic LSTMs and tensor based convolution networks. We then propose unique networks to encapsulate temporal-granularity among the modalities which is essential while extracting information within asynchronous sequences. We then integrate these two kinds of information via a fusion layer and call our novel multimodal fusion scheme as \textit{Deep-HOSeq} (Deep network with higher order Common and Unique Sequence information). The proposed \textit{Deep-HOSeq} efficiently discovers all-important information from multimodal sequences and the effectiveness of utilizing both types of information is empirically demonstrated on CMU-MOSEI and CMU-MOSI benchmark datasets. The source code of our proposed \textit{Deep-HOSeq} is and available at \url{https://github.com/sverma88/Deep-HOSeq--ICDM-2020}. 
\end{abstract}


\section{Introduction}

\begin{figure}[t]
\begin{center}
\captionsetup{justification=justified}
\includegraphics[width = 0.47\textwidth]{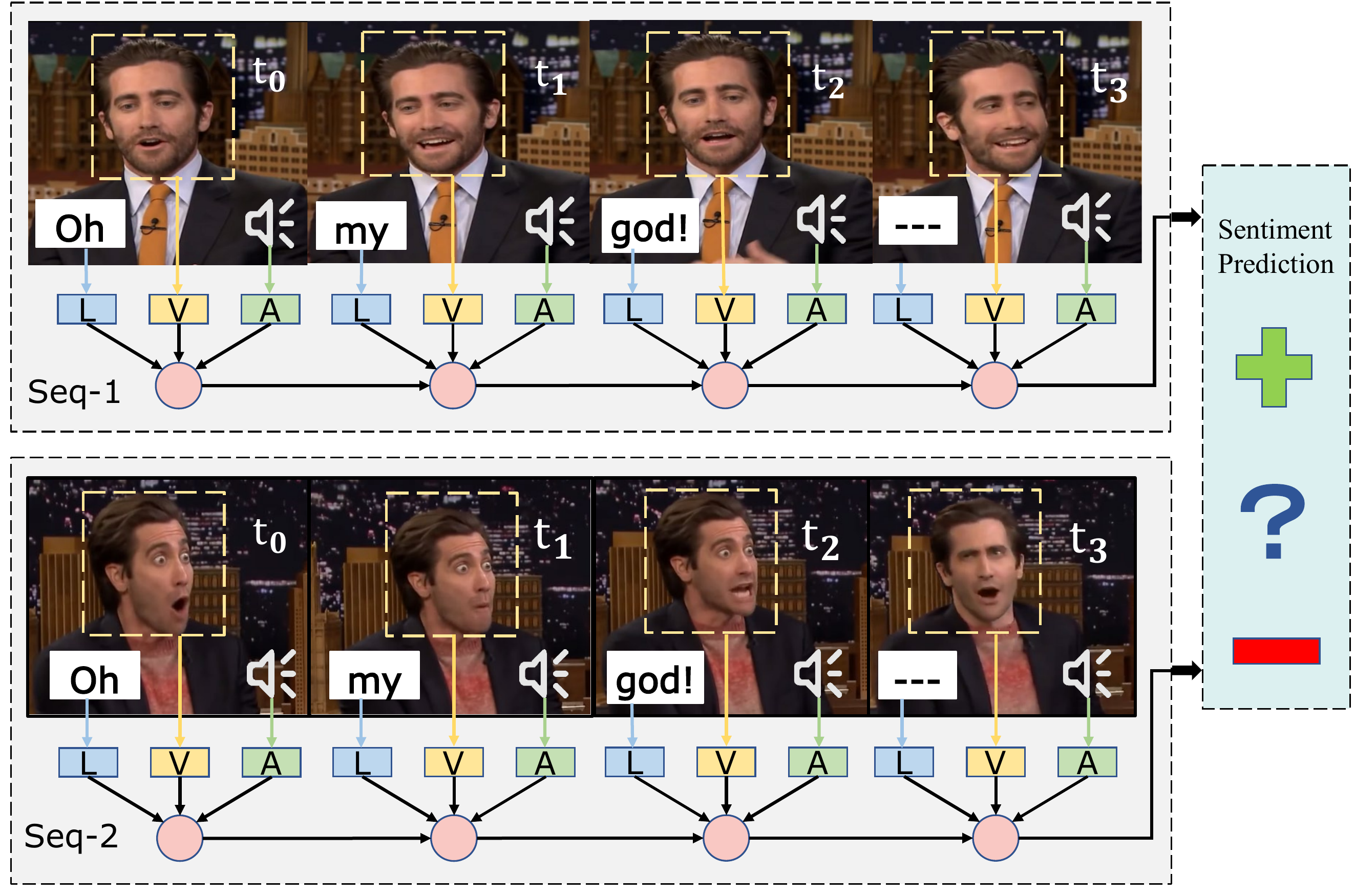}
\setlength{\belowcaptionskip}{-0.2cm}
\caption{A typical scenario illustrating different sentiments expressed with same spoken-utterance but visual gestures and vocal intonations. The asynchronous visual-gesture (occurring after the end of spoken words) at time $t_{3}$ paramountly aids in identification of the speaker's sentiment in the two sequences. Efficient processing of such asynchronous (and synchronous) temporal-interactions are a necessity for sentiment analysis through multimodal fusion.}
\label{MM}
\end{center}
\end{figure}

There is increasing popularity with sharing opinionated videos on social media platforms such as YouTube, Facebook, etc. where the speaker's sentiments are available via multiple heterogeneous forms of information such as language (spoken words), visual-gestures, and acoustic (voice).  While there has been significant development in utilizing language for sentiment analysis, a core research challenge for this domain is the efficient utilization of multimodal representations such as voice and visual gestures for sentiment prediction \cite{lahat2015multimodal,baltruvsaitis2018multimodal}. Since utilizing cues from these interacting modalities often presents a more complete view of the underlying phenomenon and thus enhances the generalization performance for sentiment prediction \cite{lahat2015multimodal,baltruvsaitis2018multimodal}. Although performing multimodal fusion for sentiment prediction is itself a challenging task due to multiple recurrent issues such as missing-values in the visual and acoustic modalities, misalignment, and etc. \cite{poria2018multimodal,Verma19}. The challenge is exacerbated when the fusion is required in the temporal domain as the multimodal temporal-interaction possesses the dual nature of promising the data-granularity and concealing its ambiguity as peril. 
A motivating example for this scenario is presented in Fig.~\ref{MM} where the speakers in both the sequences utilize the same words to express their sentiments differently. 

The speakers in both the sequences of Fig.~\ref{MM} utilize the same utterances (spoken words) to express their sentiments. Although both the sequences contain the same spoken words, the interactions between facial expressions and vocal intonations asynchronously occurring with each spoken word unveil critical information that necessitates their disparate labeling of the sequences. In particular, the facial expression at time $t_3$ drives the identification of the speaker's sentiment in the sequences. Therefore, discarding such temporal-granularity will result in the loss of critical information that substantially helps in identifying the speaker's true sentiment.  While these interactions can occur in the form of synchronous and asynchronous\footnote{Visual-Gesture occurring at the end of the spoken words.} multimodal interactions and hence, combining these temporal-cues will enhance the robustness of sentiment prediction with multimodal temporal sequences.

In this regard, to enhance the predictive power by utilizing such temporal-cues recent multimodal approaches such as MARN (Multi-Attention Recurrent Network) \cite{zadeh2018multi} and MFN (Memory Fusion Network) \cite{zadeh2018memory} combined both the inter-modal and intra-modal interaction while performing multimodal fusion. These schemes utilize series of LSTMs to obtain intra-modal dynamics and constrain them with sophisticated attention schemes (multi-attention head in MARN and delta-attention memory in MFN) to exploit the inter-modal temporal-interactions.  Although both of these techniques unanimously conclude that the utilization of both these types of information positively impacts multimodal sentiment analysis however these schemes entirely rely on the attention mechanism to discover inter-modal information and amalgamate it with intra-modal information. Their complete reliance on attention scheme is problematic due to two reasons: 1) they have deceptive attention masks (in MFN), and hence it is obscure whether the gain in prediction is attributable to inter-modal interactions \cite{pruthi2019learning} and, 2) the role of training dynamics (in MARN) instead of multiple-heads \cite{michel19neurips}. Nevertheless, both these schemes require substantial efforts to optimize the hyperparameters of their consolidated architectures to perform multimodal sequence fusion efficiently. To alleviate these drawbacks, we propose \textit{Deep-HOSeq} to perform multimodal fusion, in particular when the modalities are available as temporal sequences.

The \textit{Deep-HOSeq} performs multimodal fusion by extracting two kinds of contrasting information from multimodal temporal sequences.
The first kind of information is the amalgamation of both inter-modal and intra-modal information and can be perceived as the common\footnote{\label{note1}It should be noted that the terms `common' and `unique' information are also utilized in DeepCU \cite{Verma19} to refer to a different but related concept. The concept of common information in DeepCU is limited to inter-modal information, whereas in \textit{Deep-HOSeq}, the common information is comprised of both inter-modal and intra-modal information. Besides, the unique information in the DeepCU is comprised of factorized information from unimodality' integrated by late fusion. In contrast, the unique information in \textit{Deep-HOSeq} refers to the information present via asynchronous and synchronous temporal occurrence among modalities.}  information extracted from the modality interaction. The second type of information exploits the temporal-granularity (synchronous and asynchronous interactions within modalities, as shown in Fig.\ref{MM}) among the multimodal sequences and is derived as unique information while performing multimodal fusion. To aid the understating of proposed \textit{Deep-HOSeq} we illustrate its workflow in Fig.~\ref{DCUSeq}.

To extract these two kinds of information, we design a common network that first utilizes basic LSTM to obtain the intra-modal information from each unimodality. Then the obtained intra-modal information from each modality is amalgamated as multi-mode tensors by taking their outer-product. The elements within this multi-mode tensor reflect the strength of inter-modal interactions as correlations \cite{hou2019deep}, and this rich inter-modal information is finally captured by utilizing convolution kernels followed by fully connected layers. On the other hand, we also design a unique network for leveraging the temporal-granularity among multimodal sequences. This is achieved by first obtaining latent features from each unimodality by utilizing feed-forward layers to increase their discriminative power. We then obtain higher-order interactions within the modalities at each temporal-step followed by feature extraction with convolution layers and fully connected layers (as in the common network). We finally unify the information from all the temporal-steps with a pooling operation, which encapsulates the temporal-granularity as the unique information in \textit{Deep-HOSeq}. Although one may argue that our choice of unification scheme is not sophisticated but this scheme efficiently captures the temporal-dynamics within multimodal sequences and is demonstrated in the results section.

\begin{table}[t]
\scriptsize
\captionsetup{justification=centering}
\caption{Comparison of various multimodal fusion schemes}
\centering
\begin{tabular}{@{}l|ccccc@{}}
\toprule[1pt]
\multicolumn{1}{l}{\begin{tabular}[c]{@{}c@{}}Fusion \\ Schemes\end{tabular}}  & \multicolumn{1}{c}{\begin{tabular}[c]{@{}c@{}}Inter \\ Modal \end{tabular}}  &  \multicolumn{1}{c}{\begin{tabular}[c]{@{}c@{}}Intra \\ Modal\end{tabular}} & \multicolumn{1}{c}{\begin{tabular}[c]{@{}c@{}}Attention \\ Reliance\end{tabular}} & \multicolumn{1}{c}{Convolution} & \multicolumn{1}{c}{\begin{tabular}[c]{@{}c@{}}Multimode\\ Representation\end{tabular} }  \\ \midrule
TFN    &  \checkmark     &   $\times$      &  $\times$     &   $\times$           &   \checkmark  \\ \midrule
LMF        &  \checkmark      &   $\times$     &  $\times$    &   $\times$           &   $\times$  \\ \midrule
DeepCU    &  \checkmark      &   $\times$      &  $\times$     &   \checkmark           &   \checkmark  \\ \midrule
MFN    &  \checkmark     &   \checkmark      &  \checkmark     &   $\times$           &   $\times$  \\ \midrule
MARN        &  \checkmark     &   \checkmark     &  \checkmark    &   $\times$           &   $\times$  \\ \midrule
\textit{Deep-HOSeq} &     \checkmark   &     \checkmark    &     $\times$       &\checkmark  &  \checkmark \\ \bottomrule[1pt]
\end{tabular}
\label{comparet}
\end{table}

We finally integrate both these kinds of information with a fusion layer to perform multimodal sentiment prediction and call our novel multimodal fusion scheme as \textit{Deep-HOSeq} (Deep Higher-Order Sequence Fusion). An important characteristic of our \textit{Deep-HOSeq} is that it does not rely on attention-based schemes and hence does not face the same critiques as state of the art (SOTA) techniques such as MARN and MFN. Its superiority lies in simple but careful design choices that enable joint discovery and utilization of all-essential information to perform multimodal fusion. To aid the understanding of our technique, we summarize the similarities and differences between \textit{Deep-HOSeq} and SOTA techniques in Table.~\ref{comparet}. Besides, our major contributions in this work are summarized as below:
\begin{enumerate}
   \item We design a common network to extract both intra-modal and inter-modal information in a cascaded framework for multimodal fusion. Conceptually, the information obtained by our common network is more expressive than the SOTA as we utilize convolution on multi-mode tensors, which efficiently captures all-essential inter-modal interactions. Besides, the use of basic LSTMs efficiently discovers the underlying intra-modal dynamics and does not require strenuous efforts for parameter optimization.
    \item We design a unique network that encapsulates the temporal-granularity from multimodal sequences. This enhances the \textit{Deep-HOSeq}'s robustness with multimodal synchronous and asynchronous interactions.
   \item We design a deep consolidated network for joint discovery and utilization of both common and unique information from multimodal temporal sequences, which we call as \textit{Deep-HOSeq}.    
    \item We perform comprehensive experiments on multimodal CMU-MOSEI and CMU-MOSI datasets and demonstrate the effectiveness of utilizing both common and unique information in comparison to other techniques.   
\end{enumerate}
The rest of the paper is organized in the following sections: Sec.~\ref{PW} presents literature review of existing multimodal fusion techniques followed by details of our proposed \textit{Deep-HOSeq} in Sec.~\ref{PM}. Experimental setup and results are described in Sec.~\ref{exp} and Sec.~\ref{result}, respectively. We finally conclude our work and discuss its possible future directions in Sec.~\ref{FW}.

\section{Related Work}
\label{PW}

We focus our review on techniques performing neural-based fusion of multimodal sequences where arguably the simplest deep architecture performing fusion of heterogeneous data is Deep Multimodal Fusion (DMF) \cite{nojavanasghari2016deep}. The DMF is a successor to the Early Fusion (EF) \cite{morency2011towards}, which is one of the most utilized non-neural technique performing multimodal data fusion. The DMF is developed to perform both a) EF: combine raw (or latent) features by concatenating them; b) late fusion: process each modality with a deep network and then synthesize their decisions. Although powerful, the DMF (and EF) is a basic technique and assumes that a modality (for example, visual) does not share any relevant information within itself. In other words, it can not leverage the intra-modal information a particular modality might offer. Hence, it is limited to express only the inter-modal interactions and thus faces the same limitation as in EF \cite{zadeh2018memory}. We now review SOTA that leverages both inter-modal and intra-modal relationships while performing multimodal fusion.

\paragraph{Memory Fusion Network (MFN)} \cite{zadeh2018memory} is a recurrent model that consists of three sub-modules a) System of LSTMs to obtain intra-modal dynamics from each unimodality; b) Delta-memory Attention Network which discovers inter-modal dynamics; and c) Multi-view Gated Memory responsible for integrating the intra-modal and inter-modal dynamics. The final input for fusion in MFN comprises of concatenated intra-modal information from LSTMs and the final state of the Multi-View Gated Memory and hence can be assumed as a sophisticated EF system. Albeit powerful, the MFN relies entirely on the attention network and the Multi-View Gated Memory to obtain inter-modal dynamics while performing multimodal fusion. This complete reliance on attention mechanism is problematic as the MFN assumes synchronous inputs which is hard to achieve in real-world scenarios, and more importantly, the reliability of attention-memory to discover inter-modal interactions is questionable as shown with deceptive attention masks in  \cite{pruthi2019learning}.

\paragraph{Multi-attention Recurrent Network (MARN)} \cite{zadeh2018multi} is also a recurrent model and consists of two sub-modules a) Long-short Term Hybrid Memory (LSTHM) that amalgamates intra-modal dynamics inter-modal temporal-dynamics by explicitly augmenting LSTM with a hybrid memory; and b) Multi-attention Block (MAB) which discovers the inter-modal dynamics and successively updates the hybrid memory of LSTHMs. Similar to MFN, the MARN also completely relies on the attention scheme to obtain the inter-modal dynamics. The key difference between the two is that the earlier utilizes basic LSTMs \cite{hochreiter1997long}, whereas the latter augments a hybrid memory within the LSTMs. Besides, the MARN attributes usage of multiple-attention for gains in the predictive performance, but it is obscure whether it is from the discovery of inter-modal information or the training dynamics of the MAB (and LSTHM, which are much strenuous than MFN) \cite{michel19neurips}.

Differently from the above, few notable multimodal fusion techniques which do utilize sophisticated attention mechanisms are Tensor Fusion Networks (TFN) \cite{zadeh2017tensor}, Low-rank Multimodal Fusion (LMF) \cite{liu2018efficient}, and DeepCU \cite{Verma19}. These techniques perform multimodal fusion by utilizing the summarized information within visual (and acoustic) modality as its average. Although this leads to the loss of sequential information present in the form of visual and acoustic interactions. These techniques compensate for this information loss by modelling multiple combinations of inter-modal interactions, either as tensors or its low-rank factorized representation. 

Our proposed \textit{Deep-HOSeq} is similar to the above as it also aims to exploit both the inter-modality and intra-modality relationship while performing multimodal fusion, but substantially differs from them due to the following: 
\renewcommand{\labelenumi}{\Roman{enumi}.}
\begin{enumerate}
    \item The common network in \textit{Deep-HOSeq} extracts information from inter-modal tensors obtained via modelling the intra-modal information in multimodal sequences. Since the elements of this tensor signify the correlation strength between the fusion modalities, the information obtained is not obscure or deceptive as in MARN and MFN.
    \item Obtaining inter-modal temporal-granularity independently with unique network is a distinctive characteristic of \textit{Deep-HOSeq}, and inclusion of this information enhances the \textit{Deep-HOSeq}'s capability while dealing with asynchronous (and synchronous) multimodal sequences.   
    \item The fusion layer integrates both the common and the unique information to perform multimodal sentiment analysis. It is worth mentioning that this layer uses averaging and hence does not introduce extra model parameters. More importantly, it also refrains the common network to influence the parameters of the unique sub-network and vice-versa. This restriction allows the sub-networks to obtain complementary information and hence increase the diversity during fusion.  
\end{enumerate}

\emph{Although, all the techniques mentioned above are fundamentally different from proposed Deep-HOSeq; one must not consider the equality of Deep-HOSeq in particular to DeepCU -- based on the terms common and unique. The concept of common and unique information in both techniques is disparate and explained in detail in the footnote$^{\ref{note1}}$. Furthermore, the feature dissection process is also distinct in the technique where the earlier is proposed to perform multimodal fusion from asynchronous (and synchronous) interactions within temporal sequences whereas the latter is proposed to perform fusion of independent data units.} 

% \emph{However, all these techniques are fundamentally different from proposed \textit{Deep-HOSeq}; one must not consider the equality of DeepCU as \textit{Deep-HOSeq} based on the terms common and unique. The concept of common and unique information and also the feature dissection in both techniques are distinct and explained in detail in the footnote$^\ref{note1}$.}   

\section{Proposed Methodology}
\label{PM}

\begin{figure*}[t]
\begin{center}
\captionsetup{justification=centering}
\includegraphics[width = 0.85\textwidth]{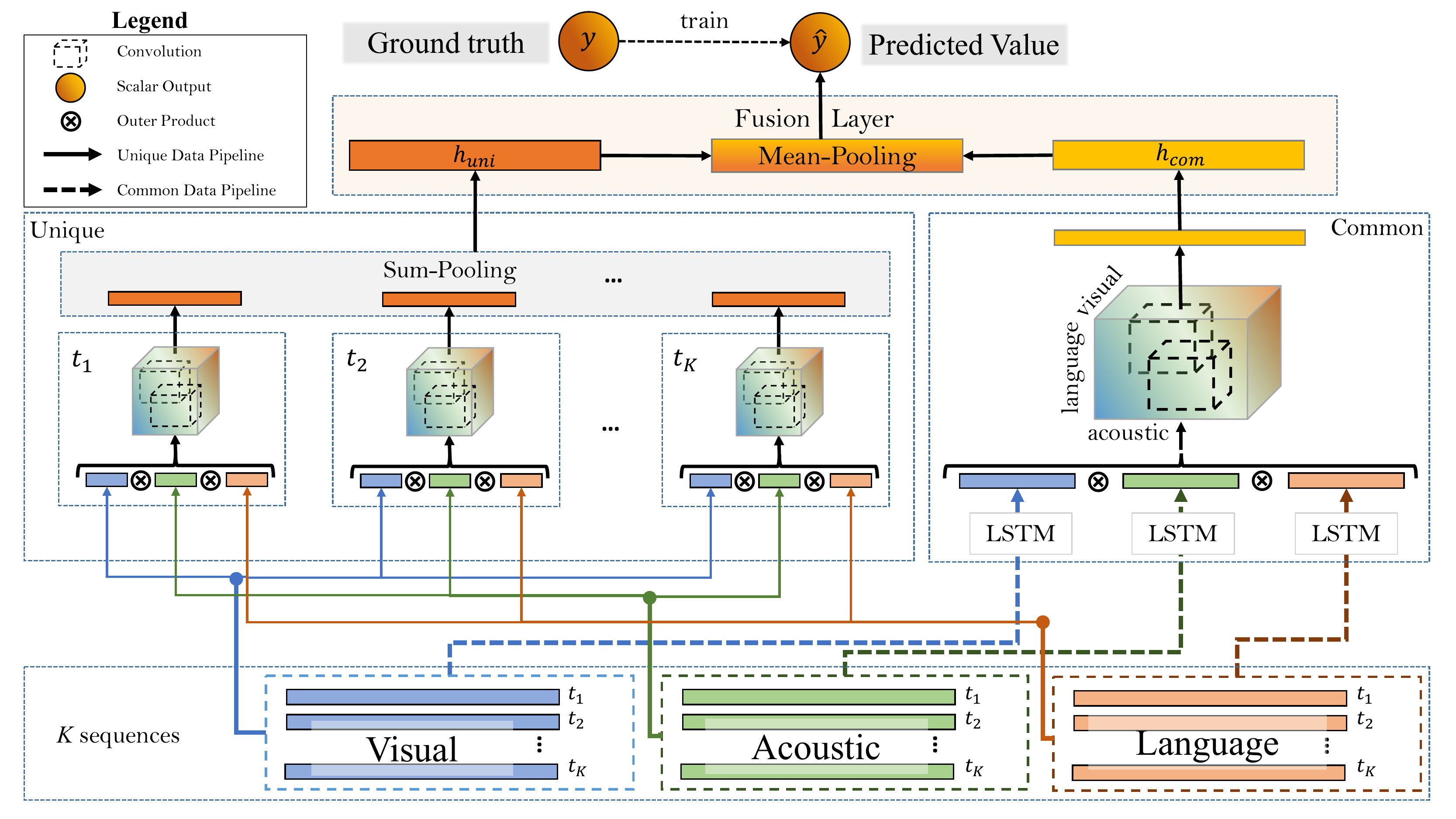}
\caption{Workflow of the proposed Deep Higher-Order Sequence Fusion network \textit{Deep-HOSeq}.}
\label{DCUSeq}
\end{center}
\end{figure*}

We aim to utilize intra-modal and inter-model dynamics by amalgamating them as common information and the dynamics of the temporal-granularity as unique information for multimodal fusion. To achieve this, we propose two sub-networks, i.e., 1) common sub-network which extracts intra-modal and inter-modal information in a cascaded manner as described in Sec.~\ref{Com}, and 2) unique sub-network for encapsulating the temporal-granularity as detailed in Sec.~\ref{Unq}. The information obtained by both networks is then integrated via a fusion layer to perform multimodal sentiment prediction in Sec.~\ref{FL}. To aid the understanding of our sub-networks, we illustrate the workflow of \textit{Deep-HOSeq} in Fig.~\ref{DCUSeq}.

We begin with raw feature vectors from a spoken utterance in the form of acoustic, visual, and language modalities denoted as $z_a \in \mathbb{R}^{1 \times t_k \times d_a}$, $z_v \in \mathbb{R}^{1 \times t_k \times d_v}$, and $z_l \in \mathbb{R}^{1 \times t_k \times d_l}$ respectively, where $d_a$, $d_v$, and $d_l$ represents the dimensionality of the feature vectors and $t_k$ represents the sequence length. These feature vectors are then independently processed with basic LSTMs in the common network and with feed-forward layers in the unique network which constrains both the sub-networks to obtain unshared latent representation. This restriction allows the sub-networks to obtain complementary feature representations at lower layers as the latent space of unique sub-network remains unaffected by the gradient from the common sub-network and vice-versa for the common sub-network. Furthermore, optimizing unshared latent space also enhances the expressiveness of \textit{Deep-HOSeq} and is empirically shown beneficial in the works \cite{he2017neural,lu2018learning}. 

% We begin with raw feature vectors from a spoken utterance in the form of acoustic, visual, and language modalities denoted as $z_a \in \mathbb{R}^{1 \times t_k \times d_a}$, $z_v \in \mathbb{R}^{1 \times t_k \times d_v}$, and $z_l \in \mathbb{R}^{1 \times t_k \times d_l}$ respectively, where $d_a$, $d_v$, and $d_l$ represents the dimensionality of the feature vectors and $t_k$ represents the sequence length. These feature vectors are then independently processed with basic LSTMs in the common network and with feed-forward layers in the unique network. This restriction of obtaining unshared latent representations allows the sub-networks to obtain complementary feature representations at lower layers as the latent space of unique sub-network remains unaffected by the gradient from the common sub-network and vice-versa for the common sub-network. Besides, optimizing unshared latent space also enhances the expressiveness of \textit{Deep-HOSeq} and is empirically shown beneficial in the works \cite{he2017neural,lu2018learning}. 

\subsection{Common Network} \label{Com}
The common network first obtains intra-modal dynamics from the individual modality by processing them with basic uni-directional LSTMs; we chose basic LSTMs as they are simple-yet-powerful enough to discover the relevant parts within a modality. We then process the latent features obtained with all the unimodal LSTMs (the final state of LSTMs) with fully-connected layers to increase their discriminatory strength, followed by an outer product to obtain multi-mode tensors. These tensors represent the amalgamated intra-modal and inter-model information within the multimodal sequences. 
\begin{equation} \label{comhid}
% \footnotesize
\begin{split}
     h_{V} &  = \sigma\Big(  LSTM \big(\boldsymbol{z}_v \big) \times W_{v} + \boldsymbol{b}_{v} \Big) \\
     h_{A} & = \sigma\Big(  LSTM \big(\boldsymbol{z}_a \big) \times W_{a} + \boldsymbol{b}_{a} \Big) \\
     h_{L} & = \sigma\Big(  LSTM \big(\boldsymbol{z}_l \big) \times W_{l} + \boldsymbol{b}_{l} \Big) \\
     T_{VAL} & = h_{V} \otimes h_{A} \otimes h_{L} \\
\end{split} 
\end{equation}

The elements of tensor $T_{VAL}$ signifies the inter-modal interactions as correlation strengths \cite{hou2019deep}, which can be efficiently derived by processing the tensor with a series of convolution and fully-connected layers as detailed in (\ref{comconv}). The highly discriminative information available after dissecting $T_{VAL}$ denoted as $h_{com}$ represents the amalgamated intra-modal and inter-modal information perceived as the common information in \textit{Deep-HOSeq}. Although this common information i.e. $h_{com}$ can still be utilized to perform multimodal prediction as in (\ref{compred}); it is still not as effective as utilizing both common and unique information, and a relative comparison of utilizing both common and information vs. only common information is presented in the experiments section.  It should be noted that our common network can be extended with sophisticated convolutions schemes such as ResNet \cite{he2016deep} etc. to boost its generalization power.
\begin{equation} \label{comconv}
% \footnotesize
\begin{split}
    \boldsymbol{\mathscr{G}}_{VAL}  & = \sigma\Big( Conv\big(T_{VAL} \big) \Big) \\
     h_{1} & = \sigma\Big({\boldsymbol{g}}_{VAL} \times W_{1} + \boldsymbol{b}_{1} \Big)  \\
     h_{2} & = \sigma\Big(h_{1} \times W_{2} + \boldsymbol{b}_{2} \Big)  \\
    & \qquad \qquad  ...\\
     h_{n} & = \sigma\Big(h_{(n-1)} \times W_{(n-1)} + \boldsymbol{b}_{(n-1)} \Big) \\
     h_{com} & =  \sigma\Big(h_{(n)} \times W_{(com)} + \boldsymbol{b}_{com} \Big) 
\end{split} 
\end{equation}
where $\boldsymbol{g}_{VAL}$ is obtained by flattening $\boldsymbol{G}_{VAL}$ to process the convolution output with fully connected layers. Multimodal sentiment prediction with common information can be obtained by utilizing $h_{com}$ as in (\ref{compred}).
\begin{equation}\label{compred}
% \footnotesize
 \hat{y}_{com} =  \big( \boldsymbol{h}_{n} \times  \boldsymbol{w}_{0_{c}} \big) + b_{0_{c}} 
\end{equation}

\subsection{Unique Network} \label{Unq}
% The unique network first processes the raw unimodal representations with a sequence of feed-forward layer to obtain their latent representations. These discriminative representations are then utilized to capture cross-categorical correlations as multi-mode tensor.  These tensors can now be processed with convolution and fully-connected layers (as in common network) to accommodate complex dynamical factors such as inter-region spatial correlations occurring between temporal sequence. This feature extraction process can be mathematically described as in (\ref{unilatent}), where $k=1,2,...,t_{k}$ i.e. the sequence length.
The unique network first obtains latent representations of raw unimodal features by processing with a sequence of feed-forward layers and then utilizes the obtained discriminative representations to capture cross-categorical correlations in the form of multi-mode tensors.  These tensors accommodate complex dynamical factors such as inter-region spatial correlations between the temporal sequence. Thus we process them with convolution and fully-connected layers (as in the common network) to extract the concealed unique information in them. The whole feature extraction process is mathematically described as in (\ref{unilatent}), where $k=1,2,...,t_{k}$ is the sequence length.
\begin{equation} \label{unilatent}
% \footnotesize
\begin{split}
     h_{V_{k}} & = \sigma\Big(z_{v_{k}} \times W_{v_{k}} + \boldsymbol{b}_{v_{k}} \Big) \\ 
     h_{A_{k}} & = \sigma\Big(z_{a_{k}} \times W_{a_{k}} + \boldsymbol{b}_{a_{k}} \Big) \\
     h_{L_{k}} & = \sigma\Big(z_{l_{k}} \times W_{l_{k}} + \boldsymbol{b}_{l_{k}} \Big) \\
     T_{VAL_{k}} & = h_{V_{k}} \otimes h_{A_{k}} \otimes h_{L_{k}} \\ 
     \boldsymbol{h}_{k} & = \sigma\Bigg( \sigma\Big( Conv\big(T_{VAL_{k}} \big) \Big) \times W_{val_{k}} + \boldsymbol{b}_{val_{k}} \Bigg) 
\end{split} 
\end{equation}
The discovery of such complex factors is essential as they promote collaboration among temporal and semantic views of the data and thus enhance classifiers' predictive performance by learning the dynamical relationship within multiple modalities \cite{huang2019mist,wu2019neural}. We finally unify the sequential information at each temporal step with a pooling operation in (\ref{unipool}). We arguably again utilize a simple-yet-effective operation to summarize the most discriminate pattern from the intermediate layer. Importantly, this unification does not add any extra parameters and provides further opportunities to increase the discriminatory strength of the unique features.
\begin{equation} \label{unipool}
\begin{split}
     h_{pool} & =  \sum_{k=1}^{t_{k}} h_{k} \\
     h_{uni} & =  \sigma\Big( h_{pool} \times W_{pool} + \boldsymbol{b}_{pool} \Big) \\
   \end{split}
\end{equation}
These pooled representations denoted as $h_{pool}$ encapsulates the temporal-granularity from the modalities interactions and are perceived as the unique information in this paper. Similar to common information, one can also utilize unique information to perform multimodal sentiment prediction, as in (\ref{uniy}).
\begin{equation} \label{uniy}
\begin{split}
    \hat{y}_{uni} & =  \big( \boldsymbol{h}_{uni} \times  \boldsymbol{w}_{0_{u}} \big) + b_{0_{u}}\\
\end{split}
\end{equation}

\subsection{Fusion Layer} \label{FL}
The feature vectors from the last hidden layer of common and the unique sub-networks are integrated by performing a mean-pooling operation followed by feed-forward layer to derive the combined prediction with common and unique information in \textit{Deep-HOSeq} in (\ref{yhat}). Our motivation for applying mean-pooling instead of any other pooling (to integrate common and unique information) is that the mean-pooling will enforce equal learning within both common and unique sub-networks as it allows equal gradient flow in both the sub-networks while training. This scenario, however, is not guaranteed in case max-Pooling (or any other pooling layer) is employed, as there is a possibility that a single network might be dominant while training and thus will result in the absence of either kind of information. The validity of this hypothesis is shown in Fig.~\ref{Q1} by comparing the performance of different kinds of information.
\begin{equation} \label{yhat}
% \footnotesize
\begin{split}
     h_{combined} & =  \textit{average\_pooling}(h_{com},h_{uni}) \\
    \hat{y}(x) & =  h_{combined} \times W + \boldsymbol{b}  \\
   \end{split}
\end{equation}
The weights of our proposed \textit{Deep-CUSeq} is optimized via  minimizing the mean square error (MSE) loss in (\ref{loss}), where $\chi$ denotes the set of multimodal training data instances, $y(x)$ denotes the target of instance $x$, and $\hat{y}(x)$ denotes the prediction obtained from \textit{Deep-HOSeq}.
\begin{equation} \label{loss}
L = \frac{1}{n} \sum_{\forall x \in \chi} ( {\hat{y}(x) - y(x)} )^2  
\end{equation}

\subsection{Complexity Analysis}
The paramount computation complexity in \textit{Deep-HOSeq} arises with obtaining unshared latent features in the unique network. This is because we utilize feed-forward layers for obtaining latent features from each temporal-sequence within modalities, and this accumulates to approximately $43\%$ of the total trainable neurons in \textit{Deep-HOSeq}. 

While a direct comparison of running time is not possible as the baselines are customised in Pytorch and \textit{Deep-HOSeq} is written in Tensorflow, but the number of parameters in the optimized MFN model is equal to $2.34 \times 10^{5}$, and for MARN it is equal to $4.58 \times 10^{5}$. Whereas, the number of trainable parameters in optimized \textit{Deep-HOSeq} is equal to $2.61 \times 10^{5}$. Additionally, both MFN and \textit{Deep-HOSeq} took less than 30 epochs to converge on the CMU-MOSEI dataset while the MARN did not converge with 1000 epochs.

\section{Experimental Setup} \label{exp}

\begin{table}[t]
\centering
\captionsetup{justification=centering}
\caption{The speaker independent splits for training, validation, and test sets from CMU-MOSEI and CMU-MOSI datasets.}
\begin{tabular}{@{}ccc@{}}
\toprule[1pt]
\textbf{Dataset} & CMU-MOSI & CMU-MOSEI \\ \midrule
\#Training Instances         & 1284     &     15290      \\
\#Validation Instances        & 229      &     2291      \\
\#Testing Instances         & 686      &    4832  \\ \bottomrule[1pt]
\end{tabular}
\label{data}
\end{table}

\subsection{Dataset.}
We perform experiments on the CMU-MOSI \cite{zadeh2016multimodal} and CMU-MOSEI datasets \cite{zadeh2018multimodal} where both the datasets consist of opinion videos collected from YouTube with only a single person in front of the camera expressing his opinion. The CMU-MOSI dataset consists of reviews from $93$ distinct speakers where each video consists of multiple opinion segments with a total of $2199$ utterances (segments) in the whole dataset. On the other hand, the CMU-MOSEI dataset consists of reviews from $1000$ distinct speakers with a total of $23,453$ utterances. Each utterance in the video is annotated with the sentiment in the range $[-3,\ 3]$. Here $-3$ indicates highly negative and $+3$ indicates highly positive sentiment. 

\subsubsection{Features}
We accessed the language, visual, and acoustic features provided by the authors \cite{zadeh2016multimodal} at their official publicly available repository\footnote{\url{https://github.com/A2Zadeh/CMU-MultimodalSDK}, SDK Version 1.0.1}. The modality specific features are provided after word alignment using P2FA \cite{yuan2008speaker} aligning them at the word granularity.  

\paragraph{Language} Pre-trained 300-dimensional Glove word embeddings \cite{pennington2014glove} were utilized to encode each sequence of transcribed word into a sequence of word vectors.

\paragraph{Visual} The library Facet\footnote{\url{https://imotions.com/}} is used to extract visual features for each frame (sampled at 30Hz). Extracted features consists of 20 facial action units, 68 facial landmarks, head pose estimates, gaze tracking and HOG features \cite{zhu2006fast}.

\paragraph{Acoustic} COVAREP acoustic framework \cite{degottex2014covarep} is utilized to extract features including 12 MFCCs, pitch, glottal source, peak, slope, voiced/unvoiced segmentation, and maxima dispersion quotient. 

The feature vectors for each modality is publicly available via CMU-MultimodalDataSDK. Also, in order to evaluate the generalization capability of models the training, testing, and validation splits of datasets are speaker-independent and pre-defined in the CMU-MultimodalDataSDK and the number of instances for both datasets are reported in Table.~\ref{data}.

\subsection{Baselines.} 
We extensively evaluate the performance of our proposed \textit{Deep-HOSeq} against neural-based and non-neural based schemes available for multimodal sentiment analysis. Thus we trained our \textit{Deep-HOSeq} and also the baselines with MSE loss in Eq.~\ref{loss}.  The details of the baselines are described as below:

\subsubsection{Early Fusion, Non-Neural Approaches}
We first collapsed the sequence dimension in all the modalities by taking their average and then trained them for a regression task by concatenating the average features. The baselines thus reported are  Support Vector Machines (SVM) and Random Forest (RF).

\subsubsection{Joint Representation, Neural Approaches}
For baselines under joint representation, we followed the protocol as in \cite{zadeh2018multimodal} and thus trained basic LSTMs on individual modalities and treated their final state as the latent features available for fusion. We then concatenated these features and trained a deep network for regression reported as EF$_{LSTM}$\cite{perez2013utterance}. We also trained RF and SVM on these latent features reported as RF-MD\cite{zadeh2016multimodal} and SVM-MD\cite{zadeh2016multimodal}, respectively. We also trained an extreme learning machine (ELM) classifier on these latent features as utilized to predict multimodal sentiment in \cite{poria2016fusing}.

\subsubsection{Deep Networks without temporal information}
As a requirement for baselines under this such as TFN \cite{zadeh2017tensor}, LMF \cite{liu2018efficient}, and DeepCU \cite{Verma19}, we collapsed the sequence dimension of acoustic and visual modalities and performed a grid search to optimize to network's hyperparameters.

\subsubsection{State of the art deep neural networks} Under this we have two state of the art techniques performing multimodal fusion with temporal sequences.

MARN (Multi Attention Recurrent Networks) \textit{SOTA1} \cite{zadeh2018multi} is described in Sec.~\ref{PW}. The source code of MARN is publicly available and a grid search is performed to optimize the network's hyperparameters on MOSEI dataset. 

MFN (Memory Fusion Networks) \textit{SOTA2}  \cite{zadeh2018memory} is another state of the art as described in Sec.~\ref{PW}. The source code is publicly available and a grid search is performed to optimize the network's hyperparameters on MOSEI dataset. 

\subsection{Parameter Setting in Deep-HOSeq} \label{PS}
We implemented \textit{Deep-HOSeq} in TensorFlow\footnote{Link to our \textit{Deep-HOSeq}'s source code repository is avaliable at: \url{https://github.com/sverma88/Deep-HOSeq--ICDM-2020}} and optimized it by minimizing the loss in Eq.~\ref{loss} with Adam Optimizer \cite{tieleman2012lecture}. The learning rate was set to $6 \times 10^{-3}$ with a mini batch size of $256$. To avoid over-fitting we applied dropout \cite{srivastava2014dropout} in our model and tune the dropout probability from [0.05, 0.8] with a step size of 0.05. The optimal dimensions of latent space in each sub-network was searched in $[5,10,15,20,25,30]$, while the number of convolution filters were set between $[1,3]$. We also applied batch-normalization \cite{ioffe2015batch} to the convolution layers to speed up the training of \textit{Deep-HOSeq}. Besides, we utilized basic uni-directional LSTM-cell \footnote{\url{https://www.tensorflow.org/api_docs/python/tf/nn/rnn_cell/BasicLSTMCell}} for obtaining the intra-modal dynamics. Lastly, we employed early stopping as in \cite{zadeh2018memory}, where the training is terminated if the MAE on the validation-set did not improved in 5 consecutive epochs.

\subsection{Evaluation Metrics}
We evaluate the performance of the baselines and \textit{Deep-HOSeq} for regression, binary classification (positive and negative sentiments),  and multi-class classification (7 sentiments). In this regard, we report Mean Absolute Error (MAE) and Pearson's Correlation (Correlation) for regression, and in the case of binary classification we report accuracy and F1-score. Whereas for multi-class classification we only report accuracy. Note that, for all metrics, higher value is better except for MAE where a lower value is better.  

Besides, while calculating the binary and multi-class accuracies, we followed the protocol in \cite{liu2018efficient,Verma19} and map the predicted sentiment (and the true sentiment) to integer values.

\section{Results and Discussions} \label{result}

The key contribution of this work is utilization of both common and unique information for multimodal data fusion. Therefore, in order to study the significance of proposed \textit{Deep-HOSeq} and its relative sub-components we performed the experiments as per the following research questions:

\subsection{Does the integration of both common and unique information enhance the generalization performance in \textit{Deep-HOSeq} or does it deteriorate the performance?}

\begin{figure}[t]
\begin{center}
\captionsetup{justification=centering}
\includegraphics[width = 0.41\textwidth]{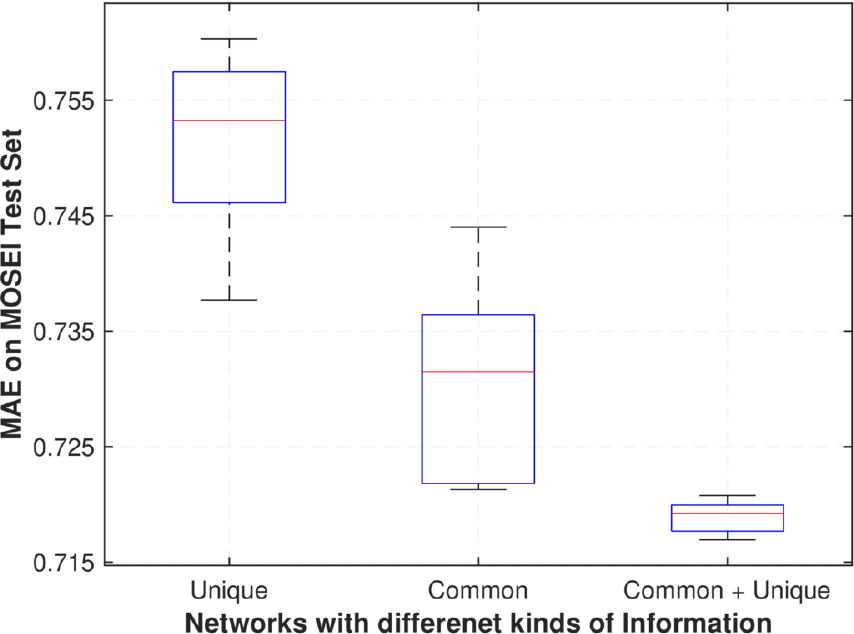}
\caption{Performance comparison of Deep-HOSeq vs. only common and unique information on the CMU-MOSEI test dataset.}
\label{Q1}
\end{center}
\end{figure}

To evaluate the effectiveness of integrating both the common and unique information in \textit{Deep-HOSeq}, we studied the performance of multimodal sentiment prediction by considering information from a) unique sub-network; b) common sub-network; and c) \textit{Deep-HOSeq}. In this regard, we obtained the performance of these different kinds of information by performing a grid search on all the hyper-parameter settings as in Sec.~\ref{PS} and present their optimized MAE on CMU-MOSEI test dataset with a box-plot in Fig.~\ref{Q1}.

First, the plot clearly suggests that the integration of both the common and the unique information is beneficial for performing multimodal sentiment analysis. We argue that this is because the \textit{Deep-HOSeq} efficiently leverages the advantages of both the amalgamated inter-modal and inter-modal information as common information and temporal-dynamics as unique information.

Second, the plot also shows that the common information achieves lower MAE compared to the unique information suggesting that the information obtained by amalgamating the intra-modal and inter-modal dynamics is more important than the temporal-granularity of multimodal sequences. Although, this might depend on the data as there may be fewer instances with high temporal-variability in the testing dataset. 

\begin{figure*}[t]
\captionsetup{justification=centering}
\begin{center}
 \includegraphics[width=1.7\columnwidth] {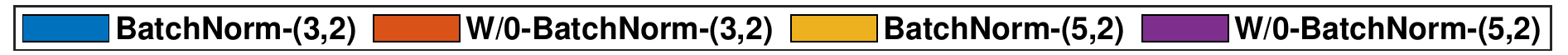}
  \vfil
    \subfloat[Analysis of activation function in lower layers of the unique sub-network.]{
 \includegraphics[width=0.42\textwidth] {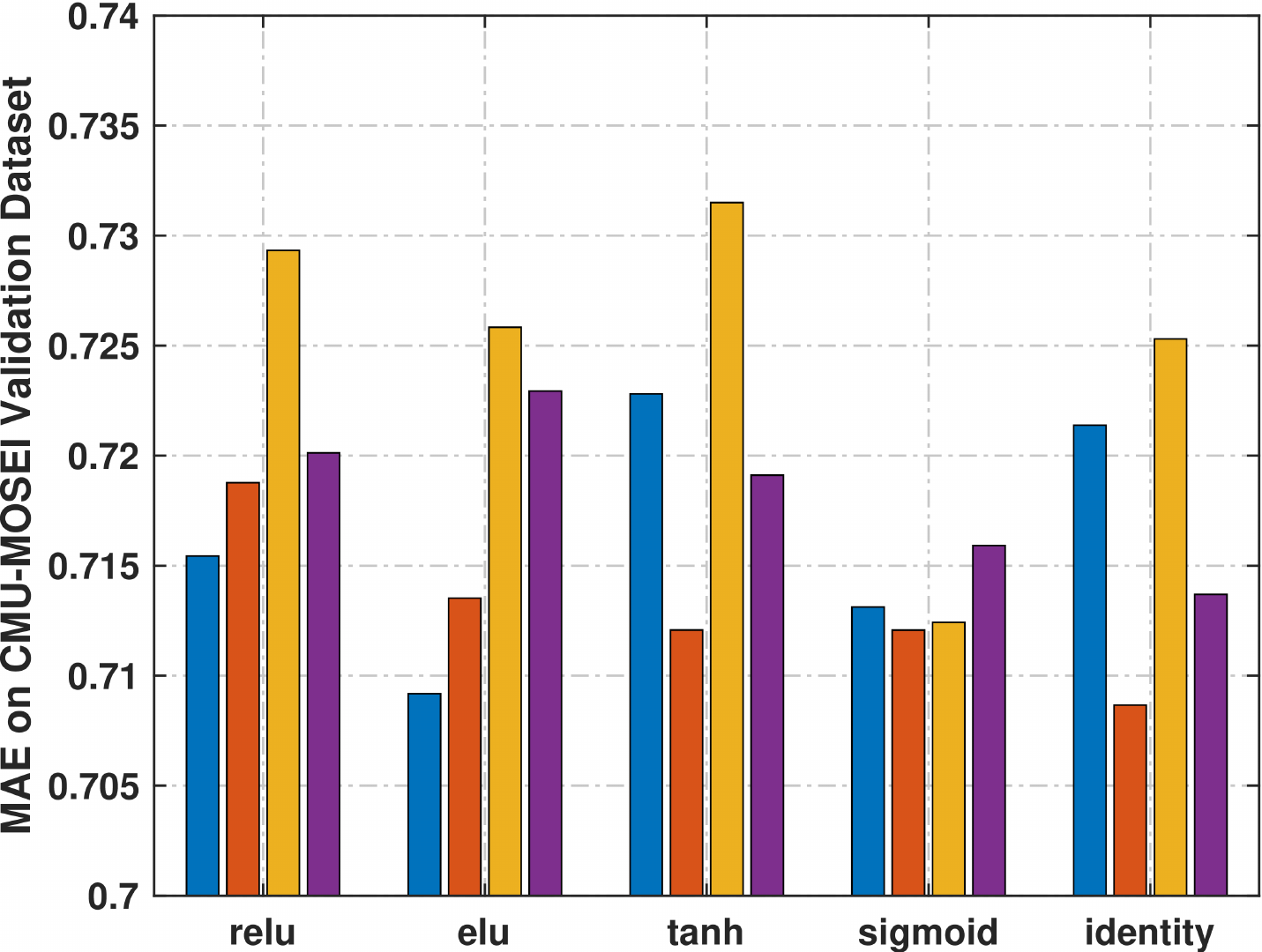}~~~
 \label{unqcomp}
}
\;
\subfloat[Analysis of activation function in lower layers of the common sub-network.]{
 \includegraphics[width=0.42\textwidth] {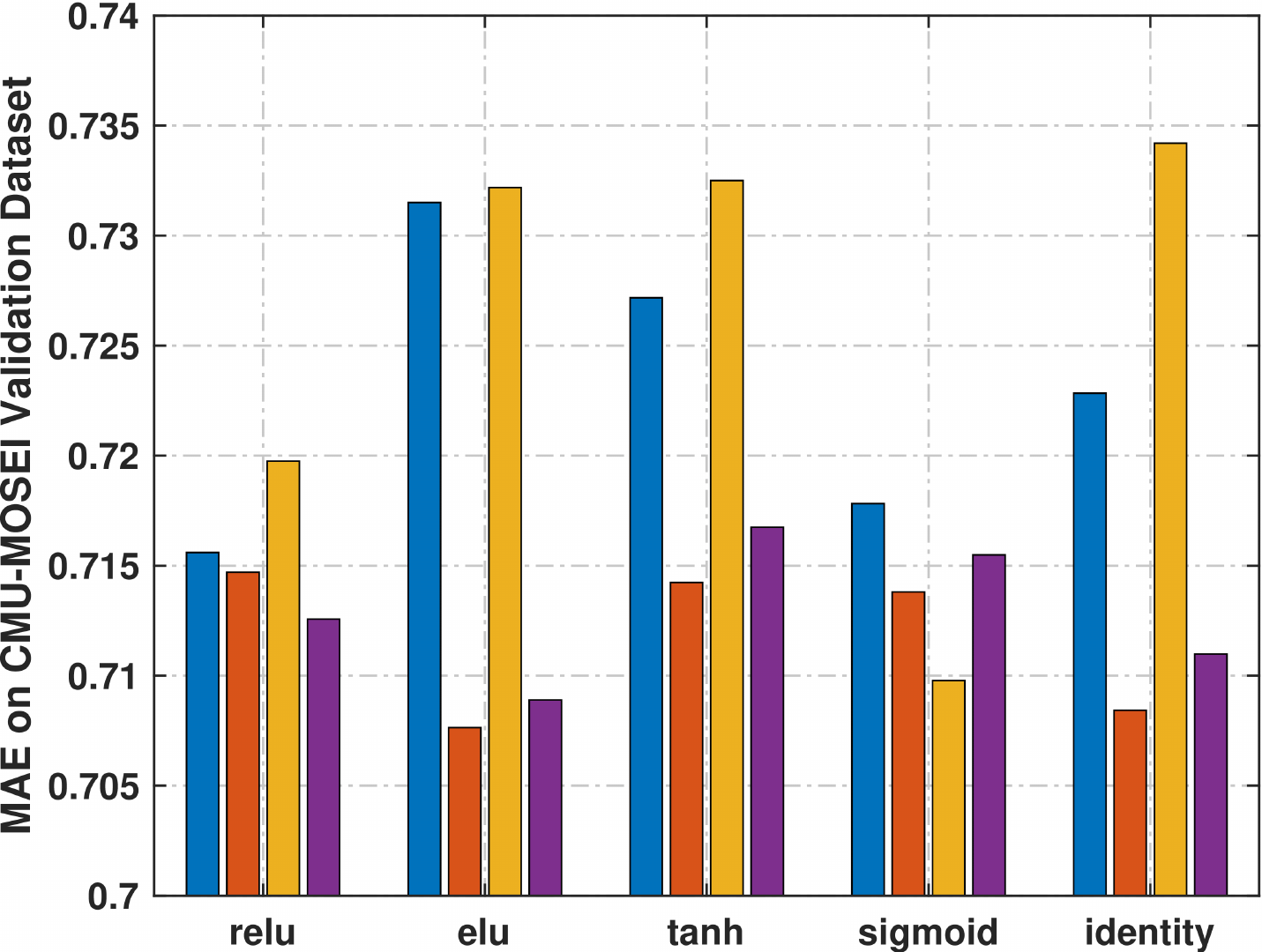}
 \label{comcomp}
}

\caption{Hyperparameter study of common and unique sub-networks on the CMU-MOSEI validation set. In the legend, integers x,y represents the kernel size and stride size in the convolution layers.}
\label{CUcomp}
\end{center}
\end{figure*}

\begin{figure}[t]
\begin{center}
\captionsetup{justification=justified}
\includegraphics[width = 0.41\textwidth]{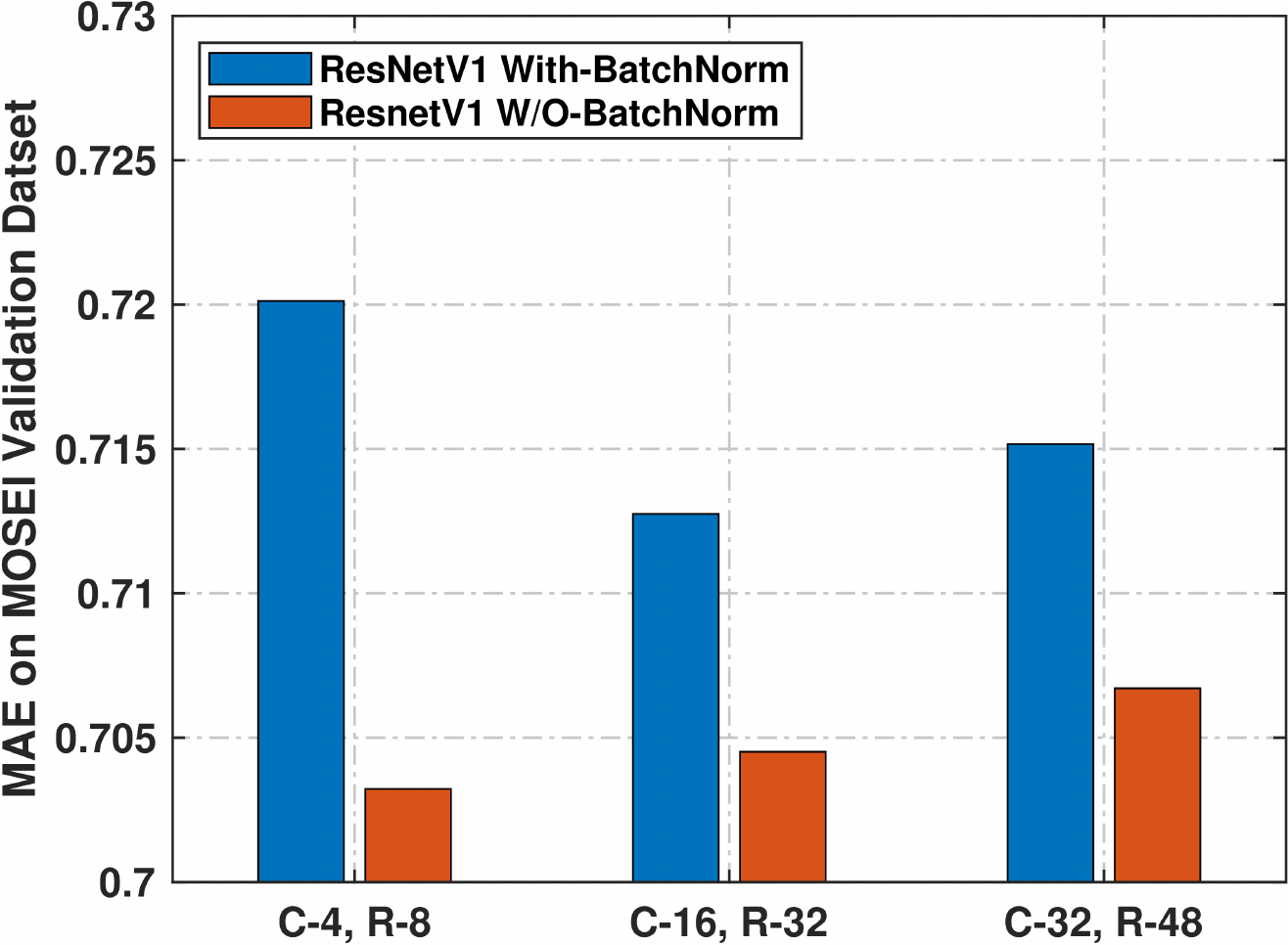}
\caption{Performance of ResNet-V1 by varying the number of convolution kernels in the common sub-network on the CMU-MOSEI dataset. The integers in the legend represents the number of kernels in the previous convolution layer and the ResNet layer respectively.}
\label{ResNet}
\end{center}
\end{figure}

\begin{figure}[t]
\begin{center}
\captionsetup{justification=justified}
\includegraphics[width = 0.41\textwidth]{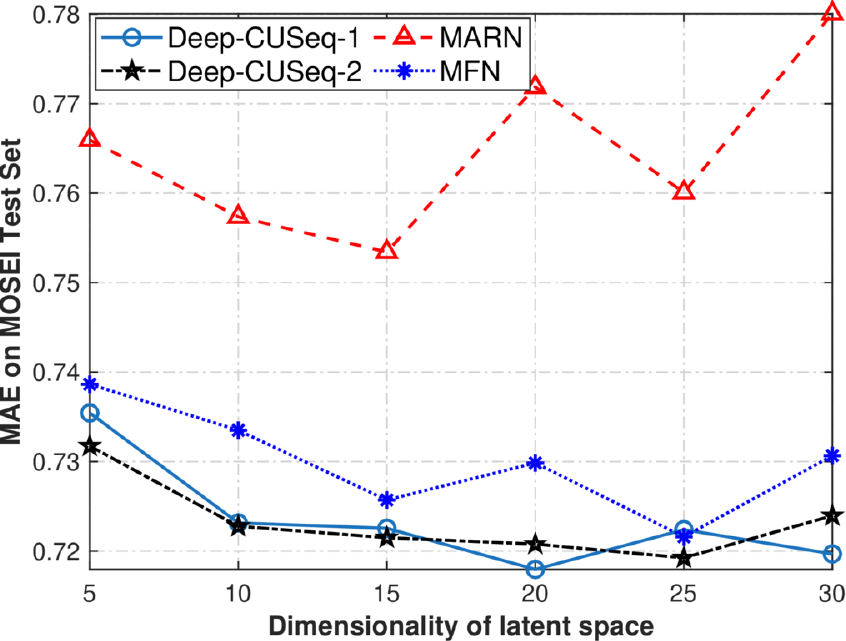}
\caption{Performance of \textit{Deep-HOSeq}, MARN, and MFN by varying hyperparameters on the CMU-MOSEI dataset. The integer x in \textit{Deep-HOSeq}-x represents the number of convolution kernels.}
\label{Q2}
\end{center}
\end{figure}

% \begin{figure}[t]
%     \centering
%   \begin{center}
% \captionsetup{justification=justified}
%         \subfloat[Analysis of activation function in lower layers of unique sub-network.]{\includegraphics[width=0.235\textwidth,height=3.4cm]{figures/Unq_MAE_noxlabel.eps}}
%         \;
%         \subfloat[Analysis of activation function lower layers of common sub-network.]{\includegraphics[width=0.235\textwidth,height=3.4cm]{figures/Com_MAE_noxlabel.eps}}
%     \label{unqcomp}
% % \setlength{\abovecaptionskip}{-0.001cm}
% % \setlength{\belowcaptionskip}{-0.4cm}
%     \caption{Hyperparameter study of unique sub-network on the CMU-MOSEI validation set. In the legend, BN and NoBN state-- with batch normalization and without batch normalization on the convolution layers. Besides, BN-x-y: x = kernel size and y = stride size.}
% \label{comcomp}
% \label{CUcomp}
% \end{center}
% \end{figure}

\subsection{How does the hyper-parameters affect the performance of unique, common, and \textit{Deep-HOSeq}?}

To answer the above question, we present a comprehensive study to understand the effects of various hyper-parameters on the performance of \textit{Deep-HOSeq}. In this regard, we split the discussion into three subsections and first discuss the effects of hyper-parameters on common and unique networks. We then discuss ways to enhance the feature discriminability of common networks by utilizing sophisticated convolution networks such as ResNet. We finally utilize these optimized hyperparameters and study the effect of varying the dimensionality of the latent space (multi-mode tensor) on \textit{Deep-HOSeq}.

\subsubsection{Analysis of common and unique sub-network}
We first analyze the effects of various hyperparameters in the common and unique subnetworks, in particular the effect of activation functions in the lower layers. Since activation functions such as \textit{relu} can lead to a sparse multi-mode tensor as compared to \textit{identity} or \textit{sigmoid} activation functions. We are thus interested in understanding whether extracting features from a full or a sparse tensor has significant effects on the networks' prediction performance. In this regard, we plot the MAE achieved by concurrently optimizing the dropout ratio and the size of convolution kernel with constant size of the latent dimensions as $10$ in the common and unique sub-networks in Fig.~\ref{CUcomp}. The \textit{x-axis} in this figure represents the choice of different activation functions in the lower layers. Besides, we have repeated the same experiment by applying batch normalization referred to as \textit{BatchNorm} and without batch normalization referred to as \textit{W/O-BatchNorm} in the legend of Fig~\ref{CUcomp}.

The performance comparison of different activation functions in the two sub-networks clearly dictates that \textit{relu} as an activation function performs equally good as any other activation function. However, we empirically found that the two networks converge faster with \textit{relu} activation than others. Moreover, these performances also suggest that applying batch normalization has a negligible effect on the performance of the sub-networks. However, empirically we found that the networks with batch normalization layer converged much faster than their counterparts. Besides, both the sub-networks perform marginally better with smaller kernel size as this might be due to the increase in overlapping regions between segments in the multi-mode tensors.

\subsubsection{Analysis of ResNet layer in common network}

The performance study of applying ResNet on the common sub-network does not yield any new insights. However, it confirms that applying batch normalization does not have a significant advantage to our model. This might be justifiable as contradictory to the image recognition the inputs to our network are features extracted by pre-trained networks. Therefore, the network does not face a covariate shift during prediction. However, the networks did converge faster with batch normalization.

We decide to drop ResNet as a layer from our model as the predictive performance of \textit{Deep-HOSeq} with a simple convolution layer is as good as with a sophisticated ResNet layer. Moreover, the number of trainable parameters increases with the ResNet layer and encourages us to proceed with a simple convolution layer in our common sub-network. 

The failure of ResNet scheme in our model might be due to less amount of training data or the depth of the network, which is at most 2 in our case.
We would like to perform some statistical analysis in the future to investigate the role of ResNet in such kind of similar schemes.

\begin{table*}[t]
\scriptsize
\captionsetup{justification=centerlast}
\caption {Performance comparison of \textit{Deep-HOSeq} vs. other fusion techniques on CMU-MOSEI dataset. Each performance metric is executed for $5$ times and their mean and standard deviation are reported. Note for all the metrics a higher value is better but for MAE.}
\centering
\begin{tabular}{@{}lccccc@{}}
\toprule[1pt]
\multicolumn{1}{c}{\multirow{2}{*}{\begin{tabular}[c]{@{}c@{}}MOSEI \\ Dataset\end{tabular}}} & \multicolumn{2}{c}{Regression} & \multicolumn{2}{c}{Binary} & 7-class \\ \cmidrule(l){2-3}  \cmidrule(l){4-5} \cmidrule(l){6-6}
\multicolumn{1}{c}{}                                                                         & MAE (lower is better)            & Correlation          & Accuracy          & F1          & Accuracy     \\ \midrule[1pt]
\textit{RF}     &       0.7794 $\pm$ \ 3.54 $\times$ $10^{-3}$ &     0.4229 $\pm$ \  7.40 $\times$ $10^{-3}$    &  70.63  $\pm$ \  3.16 $\times$ $10^{-3}$    &  70.98  $\pm$ \  3.12  $\times$ $10^{-3}$ &  39.23  $\pm$ \  3.53 $\times$ $10^{-3}$      \\
\textit{SVR}     &  0.7758 $\pm$ \ 1.66 $\times$ $10^{-3}$ &  0.4348 $\pm$ \ 1.82 \ $\times$ $10^{-3}$      &  70.53 $\pm$ \  5.96  $\times$ $10^{-4}$        &  71.06   $\pm$ \  5.39 $\times$ $10^{-4}$    &   37.64  $\pm$ \  6.68 $\times$ $10^{-3}$    \\  \midrule
\textit{EF}$_{LSTM}$ \cite{perez2013utterance}    & 0.7861 $\pm$ \ 1.03 $\times$ $10^{-2}$        &   0.3815   $\pm$ \ 3.78 $\times$ $10^{-2}$    &  71.03 $\pm$ \  2.76  $\times$ $10^{-3}$      & 71.80 $\pm$ \  9.01 $\times$ $10^{-3}$   &   40.34 $\pm$ \  5.86 $\times$ $10^{-3}$        \\
\textit{RF-MD}     &     0.7995 $\pm$ \  1.93 $\times$ $10^{-2}$  &     0.3879 $\pm$ \ 2.86 $\times$ $10^{-2}$     &  72.08 $\pm$ \  3.06 $\times$ $10^{-3}$      &  72.17 $\pm$ \ 4.76  $\times$ $10^{-3}$  &  38.81 $\pm$ \ 1.18  $\times$ $10^{-2}$        \\
\textit{SVM-MD} \cite{zadeh2016multimodal}     &  0.7886 $\pm$ \ 2.34 $\times$ $10^{-2}$ &  0.3906 $\pm$ \ 4.46  $\times$ $10^{-2}$        &  72.03 $\pm$ \  3.09  $\times$ $10^{-3}$          &  71.71 $\pm$ \ 4.74 $\times$ $10^{-3}$       &   38.87 $\pm$ \ 1.44  $\times$ $10^{-2}$ \\ 
\textit{ELM} \cite{poria2016fusing}      &   0.7699 $\pm$ \ 3.37 $\times$ $10^{-3}$   &  0.4439 $\pm$ \ 6.28 $\times$ $10^{-3}$        &  69.06 $\pm$ \ 2.04 $\times$ $10^{-3}$       &  70.41 $\pm$ \  2.19 $\times$ $10^{-3}$ &   39.62 $\pm$ \ 5.06 $\times$ $10^{-3}$ \\ \midrule 
\textit{TFN} \cite{zadeh2017tensor}  &  0.7483 $\pm$ \ 1.06 $\times$ $10^{-2}$      & 0.5005 $\pm$ \ 6.62 $\times$ $10^{-3}$  &  69.08 $\pm$ \ 1.74 $\times$ $10^{-2}$      &  69.34 $\pm$ \ 1.08 $\times$ $10^{-2}$       &      40.88 $\pm$ \ 2.03 $\times$ $10^{-2}$        \\ 
\textit{LMF} \cite{liu2018efficient}  &  0.7417 $\pm$ \ 1.19 $\times$ $10^{-2}$      & 0.5058 $\pm$ \ 1.09 $\times$ $10^{-2}$  &  71.04 $\pm$ \ 1.04 $\times$ $10^{-2}$      &  71.32 $\pm$ \ 6.23 $\times$ $10^{-3}$       &      40.64 $\pm$ \ 1.60 $\times$ $10^{-3}$        \\ 
\textit{DeepCU} \cite{Verma19}  &  0.7331 $\pm$ \ 4.32 $\times$ $10^{-3}$    &  0.5125 $\pm$ \ 3.54 $\times$ $10^{-3}$       &  71.82 $\pm$ \  8.96 $\times$ $10^{-3}$     &   70.87 $\pm$ \ 7.50 $\times$ $10^{-3}$    &   41.30 $\pm$ \ 4.29 $\times$ $10^{-3} $       \\ \midrule 
\textit{MARN} (SOTA 1) \cite{zadeh2018multi} & 0.7532 $\pm$ \ 4.46 $\times$ $10^{-2}$         & 0.4828 $\pm$ \ 4.19 $\times$ $10^{-2}$        & 68.12 $\pm$ \ 3.03 $\times$ $10^{-2}$        & 69.10 $\pm$ \ 2.82 $\times$ $10^{-2}$   & 39.39 $\pm$ \ 8.75 $\times$ $10^{-3}$     \\ 
\textit{MFN} (SOTA 2) \cite{zadeh2018memory} & 0.7270 $\pm$ \ 7.47 $\times$ $10^{-3}$          & 0.5243 $\pm$ \  3.54 $\times$ $10^{-3}$        & 72.47 $\pm$ \ 8.09 $\times$ $10^{-3}$        & 73.11 $\pm$ \ 7.06 $\times$ $10^{-2}$       & 42.69 $\pm$ \ 3.97 $\times$ $10^{-3}$   \\ \midrule[1pt]
\textit{Deep-HOSeq} (proposed) & \textbf{0.7189} $\pm$ \ \textbf{1.15} $\times$ $10^{-3}$          & \textbf{0.5438} $\pm$ \ \textbf{2.24} $\times$ $10^{-3}$        & \textbf{74.32} $\pm$ \ \textbf{8.00} $\times$ $10^{-3}$        & \textbf{75.12} $\pm$ \ \textbf{3.35} $\times$ $10^{-2}$       & \textbf{44.17} $\pm$ \ \textbf{2.60} $\times$ $10^{-3}$   \\ \bottomrule[1pt]
\end{tabular}
\label{res-mosei}
\end{table*} 

\setlength\tabcolsep{7pt}

\setlength\tabcolsep{5pt}
\begin{table}[h]
\scriptsize
\captionsetup{justification=centering}
\caption {Performance comparison of \textit{Deep-HOSeq} vs. other fusion techniques on CMU-MOSI dataset.}
\centering
\begin{tabular}{@{}lccc@{}}
\toprule[1pt]
\multicolumn{1}{c}{\multirow{2}{*}{\begin{tabular}[c]{@{}c@{}}MOSI \\ Dataset\end{tabular}}} & \multicolumn{2}{c}{Regression} & 7-class \\ \cmidrule(l){2-3}  \cmidrule(l){4-4}
\multicolumn{1}{c}{}                                                                         & MAE (lower is better)            & Correlation           & Accuracy     \\ \midrule[1pt]
\textit{TFN}  & 1.1111 $\pm$ \ 0.0003         & 0.5341 $\pm$ \ 0.0010     & 31.98 $\pm$ \ 1.1321   \\ 
\textit{LMF}   & 1.0960 $\pm$ \ 0.0021          & 0.5455 $\pm$ \  0.0032    & 30.76 $\pm$ \ 0.0339   \\ 
\textit{DeepCU}   & 1.0595 $\pm$ \ 0.0007          & 0.5506 $\pm$ \ 0.0076      & 33.14 $\pm$ \ 0.0639   \\ \midrule 
\textit{MARN}   & 1.1215 $\pm$ \ 0.0481           & 0.5116 $\pm$ \ 0.0323   & 30.54 $\pm$ \ 0.0661   \\      
\textit{MFN}   & 1.0406 $\pm$ \ 0.0568           & 0.5461 $\pm$ \ 0.0291     & 34.14 $\pm$ \ 0.0219  \\   
\midrule[1pt]
\textit{Deep-HOSeq}  & \textbf{1.0201} $\pm$ \ \textbf{0.0218}      & \textbf{0.5676} $\pm$ \  \textbf{0.0166}   & \textbf{35.87} $\pm$ \ \textbf{0.0332}    \\  \bottomrule[1pt]
\end{tabular}
\label{res-mosi}
% \vspace{-0.2cm}
\end{table}

\subsubsection{Analysis of \textit{Deep-HOSeq}}
We now plot the mean MAE obtained by varying the dimensionality of the latent space in \textit{Deep-HOSeq}'s sub-networks shown as the $x$-axis in Fig.~\ref{Q2}, and illustrate different colors to represent the number of convolution filters. Besides, as sanity checks, we also plot the optimized MAE obtained from MARN and MFN on the same latent dimensions.  A clear trend is visible in performance curves of \textit{Deep-HOSeq} in Fig.~\ref{Q2} where the MAE improves significantly by increasing the latent dimensions with noticeable improvements beyond latent space of $5$. This might be due to the size of the convolution filter which happens to be equal to the size of the multi-mode tensors, and hence applying convolutions does not prove much beneficial to \textit{Deep-HOSeq}. However, the performance gradually improves with the increase in the latent dimension supporting the learning requirement of the convolution kernels.  

A second noticeable trend from the plot is that the performance of all the fusion schemes generally improves until the latent dimensions of $15$ and then deteriorating sporadically indicating over-fitting, in particular at latent dimension of $30$. Although, the \textit{Deep-HOSeq} does not face a significant performance degradation as compared to MARN (and MFN) and this might be due to 1) less number of parameters required by convolution kernels, and 2) the ability of convolutions to efficiently capture utmost expressiveness concealed in multi-model tensors.

\subsection{Does \textit{Deep-HOSeq} provide a better mulit-modal fusion technique compared to SOTA such as MARN and MFN? Besides, are convolutions effective in obtaining inter-modal dynamics from multi-mode tensors, and whether inclusion of this information necessary?}

To address this requirement, we compare the performance of \textit{Deep-HOSeq} and baselines on the CMU-MOSEI and CMU-MOSI datasets. The performance evaluations are reported in Table.~\ref{res-mosei} and Table.~\ref{res-mosi}, respectively. On the CMU-MOSEI dataset we improve the state of the art by \textbf{3.60}\% for correlation, \textbf{2.55}\% for binary class, and \textbf{3.46}\% on multi-class accuracy. On the CMU-MOSI datset, the \textit{Deep-HOSeq} improves the correlation by \textbf{3.94}\%, and \textbf{5.07}\% on multi-class accuracy compared to SOTA approaches. 

The above results validate our hypothesis regarding a) utilizing both the common and unique information obtained with unshared latent space; b) the use of convolutions to capture utmost expressiveness offered by multi-mode representation; and c) basic LSTMs to effectively obtain intra-modal dynamics, and d) incorporating temporal-granularity from multimodal sequences with \textit{Deep-HOSeq}.

Furthermore, we can easily see that the performance achieved with the common network in Fig.~\ref{Q1} is equally competitive to the performance achieved with state of the art MFN. This indicates that our common network is able to capture inter-modal dynamics with the utilization of convolutions efficiently. It is worth mentioning that the discovery of this information does not require any attention-mechanism whose information discovery is powerful but obscure. 

Besides, on comparing the MAE achieved with EF$_{LSTM}$ \cite{perez2013utterance} against that our proposed \textit{Deep-HOSeq}. Our proposed \textit{Deep-HOSeq} reduces the MAE by \textbf{6.48}\% as it consists of both the inter-modal and intra-modal information, while EF$_{LSTM}$ only consists of intra-modal information. This demonstrates that inter-modal interactions play a significant role in multimodal fusion and hence should not be neglected.

\section{Conclusions and Future Work} \label{FW}

In this paper, we have introduced \textit{Deep-HOSeq} to perform multimodal fusion from temporal sequences. The \textit{Deep-HOSeq} integrates two kinds of information 1) common: amalgamated inter-modal and intra-modal information from multimodal temporal sequences, and 2) unique: temporal-granularity from multimodal interactions. We then demonstrated that both these two kinds of information are essential for multimodal fusion and integrated them via a fusion layer in \textit{Deep-HOSeq}. The parameters of our consolidated \textit{Deep-HOSeq} are optimized by back-propagation on the target loss function. The superiority of the combined information obtained with \textit{Deep-HOSeq} is demonstrated by performing sentiment prediction on multiple benchmark datasets where the proposed \textit{Deep-HOSeq} outperformed state of the art and other baseline approaches.  This enhancement in \textit{Deep-HOSeq} is attributable to the expressiveness from all-types of information obtained by learning complementary information from the two sub-networks. Comprehensive experiments demonstrated the effectiveness of our proposed \textit{Deep-HOSeq} for multimodal data fusion. In the future, we plan to reduce the computational complexity of the unique sub-network by designing factorized representations. Besides, generalizing \textit{Deep-HOSeq} to perform multiple tasks is another interesting research direction for this work.

% \begin{figure*}[t]
% \captionsetup{justification=justified}
%   \begin{center}
% \captionsetup{justification=justified}
%         \subfloat[Analysis of activation function in lower layers of unique sub-network.]{\includegraphics[width=0.42\textwidth]{figures/Unq_MAE_noxlabel.eps}}
%         \;
%         \subfloat[Analysis of activation function lower layers of common sub-network.]{\includegraphics[width=0.42\textwidth]{figures/Com_MAE_noxlabel.eps}}
%     \label{unqcomp}
% % \setlength{\abovecaptionskip}{-0.001cm}
% % \setlength{\belowcaptionskip}{-0.4cm}
%     \caption{Hyperparameter study of unique sub-network on the CMU-MOSEI validation set. In the legend, BN and NoBN state-- with batch normalization and without batch normalization on the convolution layers. Besides, BN-x-y: x = kernel size and y = stride size.}
% \label{comcomp}
% \label{CUcomp}
% \end{center}
% \end{figure*}

\bibliographystyle{IEEEtran}
\bibliography{Deep_HOSeq.bib}

\begin{thebibliography}{10}
\providecommand{\url}[1]{#1}
\csname url@samestyle\endcsname
\providecommand{\newblock}{\relax}
\providecommand{\bibinfo}[2]{#2}
\providecommand{\BIBentrySTDinterwordspacing}{\spaceskip=0pt\relax}
\providecommand{\BIBentryALTinterwordstretchfactor}{4}
\providecommand{\BIBentryALTinterwordspacing}{\spaceskip=\fontdimen2\font plus
\BIBentryALTinterwordstretchfactor\fontdimen3\font minus
  \fontdimen4\font\relax}
\providecommand{\BIBforeignlanguage}[2]{{%
\expandafter\ifx\csname l@#1\endcsname\relax
\typeout{** WARNING: IEEEtran.bst: No hyphenation pattern has been}%
\typeout{** loaded for the language `#1'. Using the pattern for}%
\typeout{** the default language instead.}%
\else
\language=\csname l@#1\endcsname
\fi
#2}}
\providecommand{\BIBdecl}{\relax}
\BIBdecl

\bibitem{lahat2015multimodal}
D.~Lahat, T.~Adali, and C.~Jutten, ``Multimodal data fusion: an overview of
  methods, challenges, and prospects,'' \emph{Proceedings of the IEEE}, vol.
  103, no.~9, pp. 1449--1477, 2015.

\bibitem{baltruvsaitis2018multimodal}
T.~Baltru{\v{s}}aitis, C.~Ahuja, and L.-P. Morency, ``Multimodal machine
  learning: A survey and taxonomy,'' \emph{IEEE transactions on pattern
  analysis and machine intelligence}, vol.~41, no.~2, pp. 423--443, 2018.

\bibitem{poria2018multimodal}
S.~Poria, N.~Majumder, D.~Hazarika, E.~Cambria, A.~Gelbukh, and A.~Hussain,
  ``Multimodal sentiment analysis: Addressing key issues and setting up the
  baselines,'' \emph{IEEE Intelligent Systems}, vol.~33, no.~6, pp. 17--25,
  2018.

\bibitem{Verma19}
S.~Verma, C.~Wang, L.~Zhu, and W.~Liu, ``Deepcu: Integrating both common and
  unique latent information for multimodal sentiment analysis,'' in
  \emph{Proceedings of the 28th International Joint Conference on Artificial
  Intelligence}, 2019, pp. 3627--3634.

\bibitem{zadeh2018multi}
A.~Zadeh, P.~P. Liang, S.~Poria, P.~Vij, E.~Cambria, and L.-P. Morency,
  ``Multi-attention recurrent network for human communication comprehension,''
  in \emph{32 AAAI Conference on Artificial Intelligence}, 2018.

\bibitem{zadeh2018memory}
A.~Zadeh, P.~P. Liang, N.~Mazumder, S.~Poria, E.~Cambria, and L.-P. Morency,
  ``Memory fusion network for multi-view sequential learning,'' in \emph{32
  {AAAI} Conference on Artificial Intelligence}, 2018.

\bibitem{pruthi2019learning}
D.~Pruthi, M.~Gupta, B.~Dhingra, G.~Neubig, and Z.~C. Lipton, ``Learning to
  deceive with attention-based explanations,'' in \emph{Proceedings of the 58th
  Annual Meeting of the Association for Computational Linguistics, {ACL} 2020,
  Online, July 5-10, 2020}, 2020, pp. 4782--4793.

\bibitem{michel19neurips}
P.~Michel, O.~Levy, and G.~Neubig, ``Are sixteen heads really better than
  one?'' in \emph{Advances in Neural Information Processing Systems 32},
  H.~Wallach, H.~Larochelle, A.~Beygelzimer, F.~d\textquotesingle
  Alch\'{e}-Buc, E.~Fox, and R.~Garnett, Eds., 2019, pp. 14\,014--14\,024.

\bibitem{hou2019deep}
M.~Hou, J.~Tang, J.~Zhang, W.~Kong, and Q.~Zhao, ``Deep multimodal multilinear
  fusion with high-order polynomial pooling,'' in \emph{Advances in Neural
  Information Processing Systems}, 2019, pp. 12\,113--12\,122.

\bibitem{nojavanasghari2016deep}
B.~Nojavanasghari, D.~Gopinath, J.~Koushik, T.~Baltru{\v{s}}aitis, and L.-P.
  Morency, ``Deep multimodal fusion for persuasiveness prediction,'' in
  \emph{Proceedings of the 18th ACM International Conference on Multimodal
  Interaction}.\hskip 1em plus 0.5em minus 0.4em\relax ACM, 2016, pp. 284--288.

\bibitem{morency2011towards}
L.-P. Morency, R.~Mihalcea, and P.~Doshi, ``Towards multimodal sentiment
  analysis: Harvesting opinions from the web,'' in \emph{Proceedings of the
  13th international conference on multimodal interfaces}.\hskip 1em plus 0.5em
  minus 0.4em\relax ACM, 2011, pp. 169--176.

\bibitem{hochreiter1997long}
S.~Hochreiter and J.~Schmidhuber, ``Long short-term memory,'' \emph{Neural
  computation}, vol.~9, no.~8, pp. 1735--1780, 1997.

\bibitem{zadeh2017tensor}
A.~Zadeh, M.~Chen, S.~Poria, E.~Cambria, and L.-P. Morency, ``Tensor fusion
  network for multimodal sentiment analysis,'' in \emph{Proceedings of the 2017
  Conference on Empirical Methods in Natural Language Processing}, 2017, pp.
  1103--1114.

\bibitem{liu2018efficient}
Z.~Liu, Y.~Shen, V.~B. Lakshminarasimhan, P.~P. Liang, A.~B. Zadeh, and L.-P.
  Morency, ``Efficient low-rank multimodal fusion with modality-specific
  factors,'' in \emph{Proceedings of the 56th Annual Meeting of the Association
  for Computational Linguistics (Volume 1: Long Papers)}, 2018, pp. 2247--2256.

\bibitem{he2017neural}
X.~He and T.-S. Chua, ``Neural factorization machines for sparse predictive
  analytics,'' in \emph{Proceedings of the 40th International ACM SIGIR
  conference on Research and Development in Information Retrieval}.\hskip 1em
  plus 0.5em minus 0.4em\relax ACM, 2017, pp. 355--364.

\bibitem{lu2018learning}
C.-T. Lu, L.~He, H.~Ding, B.~Cao, and P.~S. Yu, ``Learning from multi-view
  multi-way data via structural factorization machines,'' in \emph{Proceedings
  of the 2018 World Wide Web Conference on World Wide Web}.\hskip 1em plus
  0.5em minus 0.4em\relax International World Wide Web Conferences Steering
  Committee, 2018, pp. 1593--1602.

\bibitem{he2016deep}
K.~He, X.~Zhang, S.~Ren, and J.~Sun, ``Deep residual learning for image
  recognition,'' in \emph{Proceedings of the IEEE conference on computer vision
  and pattern recognition}, 2016, pp. 770--778.

\bibitem{huang2019mist}
C.~Huang, C.~Zhang, J.~Zhao, X.~Wu, D.~Yin, and N.~Chawla, ``Mist: A multiview
  and multimodal spatial-temporal learning framework for citywide abnormal
  event forecasting,'' in \emph{The World Wide Web Conference}, 2019, pp.
  717--728.

\bibitem{wu2019neural}
X.~Wu, B.~Shi, Y.~Dong, C.~Huang, and N.~V. Chawla, ``Neural tensor
  factorization for temporal interaction learning,'' in \emph{Proceedings of
  the Twelfth ACM International Conference on Web Search and Data
  Mining}.\hskip 1em plus 0.5em minus 0.4em\relax ACM, 2019, pp. 537--545.

\bibitem{zadeh2016multimodal}
A.~Zadeh, R.~Zellers, E.~Pincus, and L.-P. Morency, ``Multimodal sentiment
  intensity analysis in videos: Facial gestures and verbal messages,''
  \emph{IEEE Intelligent Systems}, vol.~31, no.~6, pp. 82--88, 2016.

\bibitem{zadeh2018multimodal}
A.~Zadeh, P.~P. Liang, S.~Poria, E.~Cambria, and L.-P. Morency, ``Multimodal
  language analysis in the wild: Cmu-mosei dataset and interpretable dynamic
  fusion graph,'' in \emph{Proceedings of the 56th Annual Meeting of the
  Association for Computational Linguistics (Volume 1)}, 2018, pp. 2236--2246.

\bibitem{yuan2008speaker}
J.~Yuan and M.~Liberman, ``Speaker identification on the scotus corpus,''
  \emph{Journal of the Acoustical Society of America}, vol. 123, no.~5, p.
  3878, 2008.

\bibitem{pennington2014glove}
J.~Pennington, R.~Socher, and C.~Manning, ``Glove: Global vectors for word
  representation,'' in \emph{Proceedings of the 2014 conference on empirical
  methods in natural language processing (EMNLP)}, 2014, pp. 1532--1543.

\bibitem{zhu2006fast}
Q.~Zhu, M.-C. Yeh, K.-T. Cheng, and S.~Avidan, ``Fast human detection using a
  cascade of histograms of oriented gradients,'' in \emph{Computer Vision and
  Pattern Recognition, 2006 IEEE Conference on}, vol.~2.\hskip 1em plus 0.5em
  minus 0.4em\relax IEEE, 2006, pp. 1491--1498.

\bibitem{degottex2014covarep}
G.~Degottex, J.~Kane, T.~Drugman, T.~Raitio, and S.~Scherer,
  ``Covarep\textemdash a collaborative voice analysis repository for speech
  technologies,'' in \emph{Acoustics, Speech and Signal Processing (ICASSP),
  2014 IEEE International Conference on}.\hskip 1em plus 0.5em minus
  0.4em\relax IEEE, 2014, pp. 960--964.

\bibitem{perez2013utterance}
V.~P{\'e}rez-Rosas, R.~Mihalcea, and L.-P. Morency, ``Utterance-level
  multimodal sentiment analysis,'' in \emph{Proceedings of the 51st Annual
  Meeting of the Association for Computational Linguistics (Volume 1)}, 2013,
  pp. 973--982.

\bibitem{poria2016fusing}
S.~Poria, E.~Cambria, N.~Howard, G.-B. Huang, and A.~Hussain, ``Fusing audio,
  visual and textual clues for sentiment analysis from multimodal content,''
  \emph{Neurocomputing}, vol. 174, pp. 50--59, 2016.

\bibitem{tieleman2012lecture}
T.~Tieleman and G.~Hinton, ``Lecture 6.5-rmsprop: Divide the gradient by a
  running average of its recent magnitude,'' \emph{COURSERA: Neural networks
  for machine learning}, vol.~4, no.~2, pp. 26--31, 2012.

\bibitem{srivastava2014dropout}
N.~Srivastava, G.~Hinton, A.~Krizhevsky, I.~Sutskever, and R.~Salakhutdinov,
  ``Dropout: a simple way to prevent neural networks from overfitting,''
  \emph{The Journal of Machine Learning Research}, vol.~15, no.~1, pp.
  1929--1958, 2014.

\bibitem{ioffe2015batch}
S.~Ioffe and C.~Szegedy, ``Batch normalization: accelerating deep network
  training by reducing internal covariate shift,'' in \emph{Proceedings of the
  32nd International Conference on International Conference on Machine
  Learning-Volume 37}.\hskip 1em plus 0.5em minus 0.4em\relax JMLR. org, 2015,
  pp. 448--456.

\end{thebibliography}

\end{document}